\documentclass[sigconf, authorversion]{acmart}

\AtBeginDocument{%
  }

\begin{document}

\title{OrDA: Orthogonal Disentanglement of Access Habits Framework for Homepage Marketing Block Recommendations}

\author{Lingxiao Zhang}
\email{owen.zlx@antgroup.com}
\affiliation{%
  \institution{Ant Group}
  \city{Hangzhou}
  \state{Zhejiang}
  \country{China}
}

\author{Xiaobo Li}
\email{yujing.lxb@antgroup.com}
\affiliation{%
  \institution{Ant Group}
  \city{Hangzhou}
  \state{Zhejiang}
  \country{China}
}

\author{Tao Xu}
\email{tomas.xt@antgroup.com}
\affiliation{%
  \institution{Ant Group}
  \city{Hangzhou}
  \state{Zhejiang}
  \country{China}
}








\renewcommand{\shortauthors}{Lingxiao Zhang, Xiaobo Li, Tao Xu}

\begin{abstract}
  Clicks on homepage marketing blocks are driven by a dual-mechanism of content interest and access habits. However, habitual clicks often create "Pseudo-Positives" in marketing slots, where position advantage masks mediocre content quality, leading to biased recommendation ecosystems.
  
  We propose a framework called \textbf{Or}thogonal \textbf{D}isentanglement of \textbf{A}ccess habits (\textbf{OrDA}) to purify interest signals. OrDA utilizes a dual-tower structure with a gated allocation layer to adaptively route features and minimize interference. To ensure rigorous separation, we employ orthogonal regularization to constrain the latent interest and habit manifolds to be geometrically perpendicular. OrDA performs causal intervention ($do$-calculus\cite{DoCalculus}) during inference to rank items solely by purified interest scores. Empirical online evaluations on large-scale datasets demonstrate that OrDA effectively eliminates access-habit bias, outperforming state-of-the-art methods in predictive accuracy.  Online A/B test shows 5.64\% user click-through rates (UCTR) improvement on the Zhima homepage marketing block, Zhima’s rent-floor recommendation.
\end{abstract}

\ccsdesc[500]{Information systems~Retrieval models and ranking}

\keywords{Disentangled Learning, Causal Intervention, Recommender System}



\maketitle

\section{Introduction}
\begin{figure}[h]
  \centering
  \includegraphics[width=0.78\linewidth]{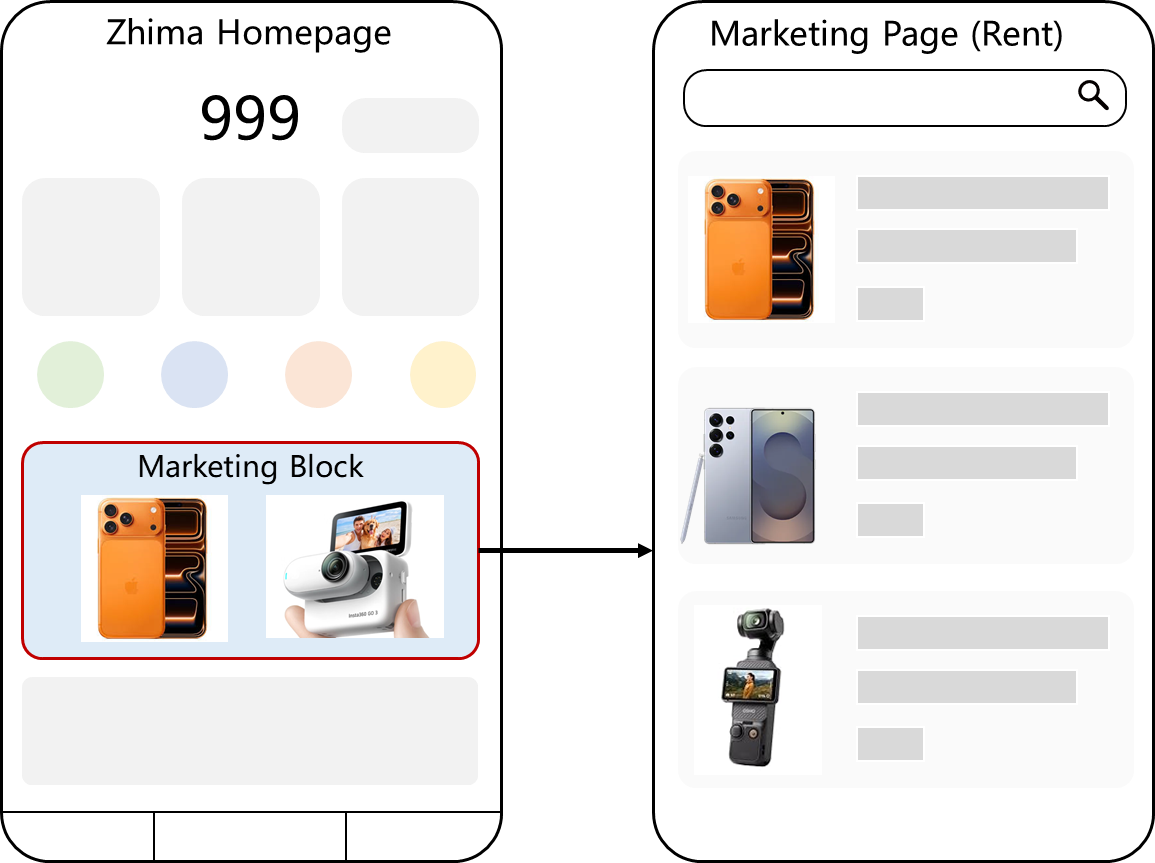}
  \caption{A demo of Zhima homepage marketing block recommendation 
  \label{fig:Homepage}
  }
  \Description{A demo of Zhima homepage marketing block recommendation}
\end{figure}

The homepage of an internet application can be summarized as having the core function of: precise navigation and ecosystem distribution from massive content to user interests. Homepage marketing blocks (shown in Fig.~\ref{fig:Homepage}) play the role of guiding the vast user traffic to different vertical sub-scenarios\cite{USD}. These blocks often suffer from a subtle yet pervasive form of confounding bias: the user-channel access habitual dependency. Specifically, in prominent marketing blocks—frequently accessed by "heavy users" as part of their routine digital journey—observed click data is often contaminated by "Pseudo-Positives." These are interactions driven not by a genuine alignment between user interests and material content, but rather by the user's habitual inertia within their primary access channels. 
Standard models, which treat all clicks as equivalent signals of preference, inadvertently capture this "habitual dividend," leading to a distorted representation of user interests. From a causal perspective, the user's channel affinity acts as a confounder, inflating the perceived utility of certain contents simply because they occupy the user's "path of least resistance." Consequently, such models often perform poorly with cold-start users and fail to generalize to broader, non-habitual contexts, as they struggle to distinguish between user-channel access habits and user-content interests.

In order to address this problem, we propose a \textbf{Or}thogonal \textbf{D}isentanglement of \textbf{A}ccess habits (\textbf{OrDA}) framework for homepage marketing block recommendations. The main contributions of this work are summarized as follows:
\begin{itemize}
\item 
\textbf{Problem Formulation of Access Habit Bias}: We provide a formal causal analysis of access habit bias prevalent in homepage marketing blocks. Unlike conventional exposure bias, we identify and define the phenomenon of habit-induced pseudo-positives, where users' navigational inertia at primary entry points masks their genuine content preferences, providing a new perspective on bias decomposition.
\item
\textbf{Orthogonal Disentanglement of Access Habits(OrDA)}: We propose OrDA framework, a novel multi-tower architecture designed to decouple user intent from habitual access patterns. Our model can effectively isolate the spurious correlations between user-content interests and user-channel access habits via orthogonal regularization and $do$-calculus-based intervention strategy.
\item
\textbf{Superior Performance on Industrial Datasets}: OrDA achieves state-of-the-art performance on Zhima homepage datasets during extensive offline experiments. Online A/B test shows 5.64\% UCTR gains for Zhima’s rent-floor recommendations.
\end{itemize}

\section{Related Work}
\subsection{Debiasing in Recommendation Systems}

The identification and elimination of biases, such as position bias\cite{PAL} and popular bias\cite{DICE}, are critical for building robust recommendation systems\cite{Bias}. Traditional approaches mainly include propensity\_based method\cite{IPW, USD}, which re-weights samples based on their exposure probability, dual learning\cite{DRJL, MultiIPW, DCMT}, which learns an auxiliary model to estimate biases, disentangled representation learning\cite{DICE, DUI}, which employs contrastive learning to separate interest from conformity, and pseudo-label method\cite{NISE}, which aims to complete the counterfactual space by assigning estimated labels to unobserved entries. Our work differs from these by specifically focusing on access habits in homepage scenarios and performing debiasing through a geometric orthogonal constraint rather than simple score adjustment to enforce a clear causal boundary and ensure a more robust representation of user intrinsic interests.

\subsection{Synergizing Multi-tower Architectures with Causal Inference}
Multi-tower models (e.g., ESMM\cite{ESMM}, MMOE\cite{MMOE}, PLE\cite{PLE}) are widely used in industrial recommendation to handle multiple tasks. In reality, these architectures provide a structural foundation for task or feature decomposition\cite{MultiIPW, DCMT, USD, ESCM2}. From a causal perspective, a significant limitation of traditional multi-tower models is feature interference: since towers often share a common embedding bottom, the gradients from a biased signal (e.g., access habit) can easily contaminate the representations of other towers (e.g., content interest). Our work advances this structural decomposition by re-interpreting the multi-tower topology as a physical realization of a Structural Causal Model (SCM).

\section{Methods}
\subsection{Problem Definition}
\begin{figure}[h]
  \centering
  \includegraphics[width=\linewidth]{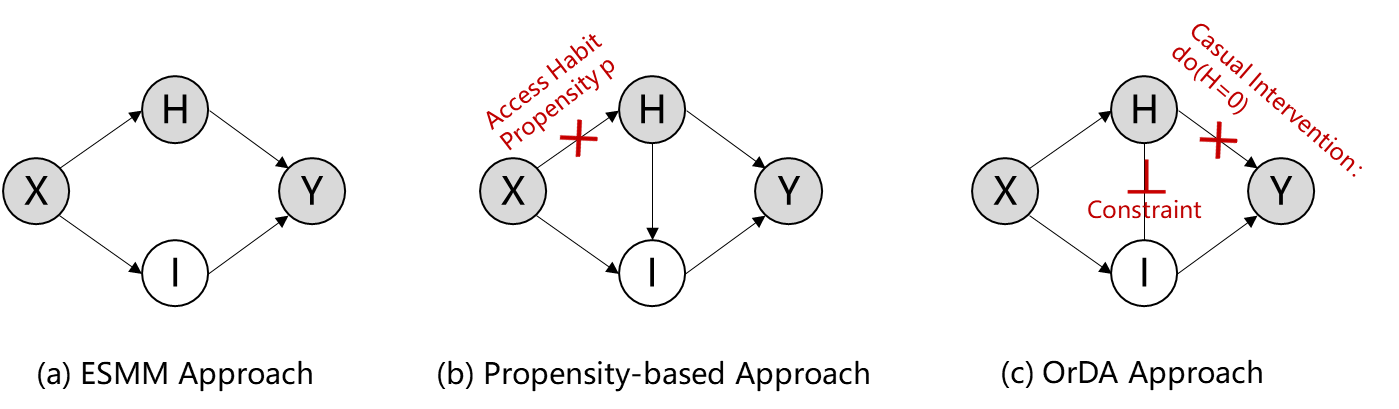}
  \caption{Causal graphs depicting training of ESMM-based Approach, Propensity-based Approach, OrDA. Hollow circles indicate latent variables, and shaded circles indicate observed variables.
  \label{fig:CasualGraph}
  }
  \Description{Causal graphs depicting training of ESMM-based Approach, Propensity-based Approach, OrDA. Hollow circles indicate latent variables, shaded circles indicate observed variables.}
\end{figure}

To formally characterize the mechanism of habit-induced pseudo-positives and justify the architecture of OrDA, we employ a causal graph to represent the data generation process in homepage marketing blocks.
As illustrated in Fig.~\ref{fig:CasualGraph}, we define the following variables:
\begin{itemize}
\item $X$: The user-content pair.
\item $H$: The latent navigational inertia or \textit{access-habit} bias.
\item $I$: The latent \textit{interest} of a user towards the content.
\item $Y$: The observed click outcome.
\item $u$: User features such as demographics preferences.
\item $c$: Content features such as category and quality.
\item $u2c$: Statistical features of a user towards the content.
\end{itemize}
We define user-channel access habit and user-content interest as the main factors governing the click behavior in homepage marketing blocks. The structural equations of these latent spaces are:
\begin{equation}
\begin{cases}
I = f_I(u, c, u2c), \\
H = f_H(u)
\label{equation1}
\end{cases}
\end{equation}
We aim to get the pure user-content interest from the confusion. Fig.~\ref{fig:CasualGraph}(a) shows ESMM-based approach\cite{ESMM} uses two towers to predict access habit probability $P(\hat{h})=\sigma(H)$ and user-content interest probability $P(\hat{i})=\sigma(I)$, and multiplies them to predict habit-induced user interest probability $P(\hat{y})=\sigma(Y)$:
\begin{equation}
P(\hat{y}) = P(\hat{h}) * P(\hat{i})
\end{equation}
This approach does not enforce a clear causal boundary or relation between two towers, and thus it introduces false independent prior. 

Fig.~\ref{fig:CasualGraph}(b) shows propensity-based approach\cite{ESCM2,DCMT,USD, MultiIPW} integrates counterfactual reasoning into multi-task learning to mitigate selection bias via inverse propensity weighting technique and $do$-calculus\cite{DoCalculus}. Habit-induced user interest probability is defined as:
\begin{equation}
P(\hat{y}) = P(\hat{h}) * P(\hat{i} | do(\hat{h}=1))
\end{equation}
During the training stage, this approach adds another tower to predict habit-induced interest probability $P(\hat{i}|do(\hat{h}=1))=\frac{P(\hat{y})}{P(\hat{h})}$. While the division form of the expression relying on numerical re-weighting may suffer from high variance and instability.

OrDA (shown in Fig.~\ref{fig:CasualGraph}(c)) extends the causal philosophy to the \textbf{representation level} as following equation: 
\begin{equation}
Y = H \oplus I
 \label{eq:equation4}
\end{equation}
where $\oplus$ denotes the joint logit fusion, representing the additive interaction between the user-content intrinsic interest and their access habit in the latent space. Our model posits that $I$ and $H$ are parallel causal drivers and a click occurs when the content satisfies the user's interest \textbf{or} the block triggers their access habits. Thus we enforce \textbf{Latent Orthogonal Constraint} to ensure that the interest tower is inherently invariant to habitual confounders.

\subsection{Training and Counterfactual Inference}
During the training stage, the loss functions of bias CTR and Habit are defined as: 
\begin{equation}
\begin{cases}
\mathcal{L}_{bCTR} = \mathcal{L}_{BCE}(y, \sigma(Y)) = \mathcal{L}_{BCE}(y, \sigma(H \oplus I)) \\ 
\mathcal{L}_{Habit} = \mathcal{L}_{BCE}(y, \sigma(H))
\end{cases}
\end{equation}
where $\mathcal{L}_{BCE}$ is Binary Cross-Entropy loss: 
\begin{equation}
 \mathcal{L}_{BCE} = -\frac{1}{N} \sum_{i=1}^{N} \left[ y_i \log(p_i) + (1 - y_i) \log(1 - p_i) \right]
 \label{eq:bce_loss}
\end{equation}
 where $y$ is the ground truth and $p$ is the probability of positive class. To ensure rigorous separation, the orthogonal cosine regularization\cite{Cosine, DomainSeparationNetworks} is employed to constrain the latent manifolds, forcing the interest and habit vectors to be geometrically perpendicular:
\begin{equation}
\mathcal{L}_{Orth} = ( \frac{\mathbf{v}_{int} \cdot \mathbf{v}_{hab}}{\|\mathbf{v}_{int}\| \cdot \|\mathbf{v}_{hab}\|} )^2
\label{equation7}
\end{equation}
where $\mathbf{v}_{int}$ is the interest tower vector and $\mathbf{v}_{hab}$ is the habit tower vector. Total loss function is as follow: 
\begin{equation}
\mathcal{L}_{total} = \mathcal{L}_{bCTR} + \alpha*\mathcal{L}_{Habit} +\beta*\mathcal{L}_{Orth}
\label{equation8}
\end{equation}
where $\alpha$ and $\beta$ are hyperparameters that control the weights of the habit task and the orthogonal task.

During the online inference stage, a causal intervention on the habit variable is performed to eliminate the pseudo-positive effect. Formally, we compute the purified user-content interest score $S$ by applying $do$-operator $do(H=0)$:
\begin{equation}
S = \mathbb{E}[Y \mid I, do(H=0)] = \sigma(I)
 \label{equation9}
\end{equation}

\subsection{Model Architecture}
\begin{figure}[h]
  \centering
  \includegraphics[width=\linewidth]{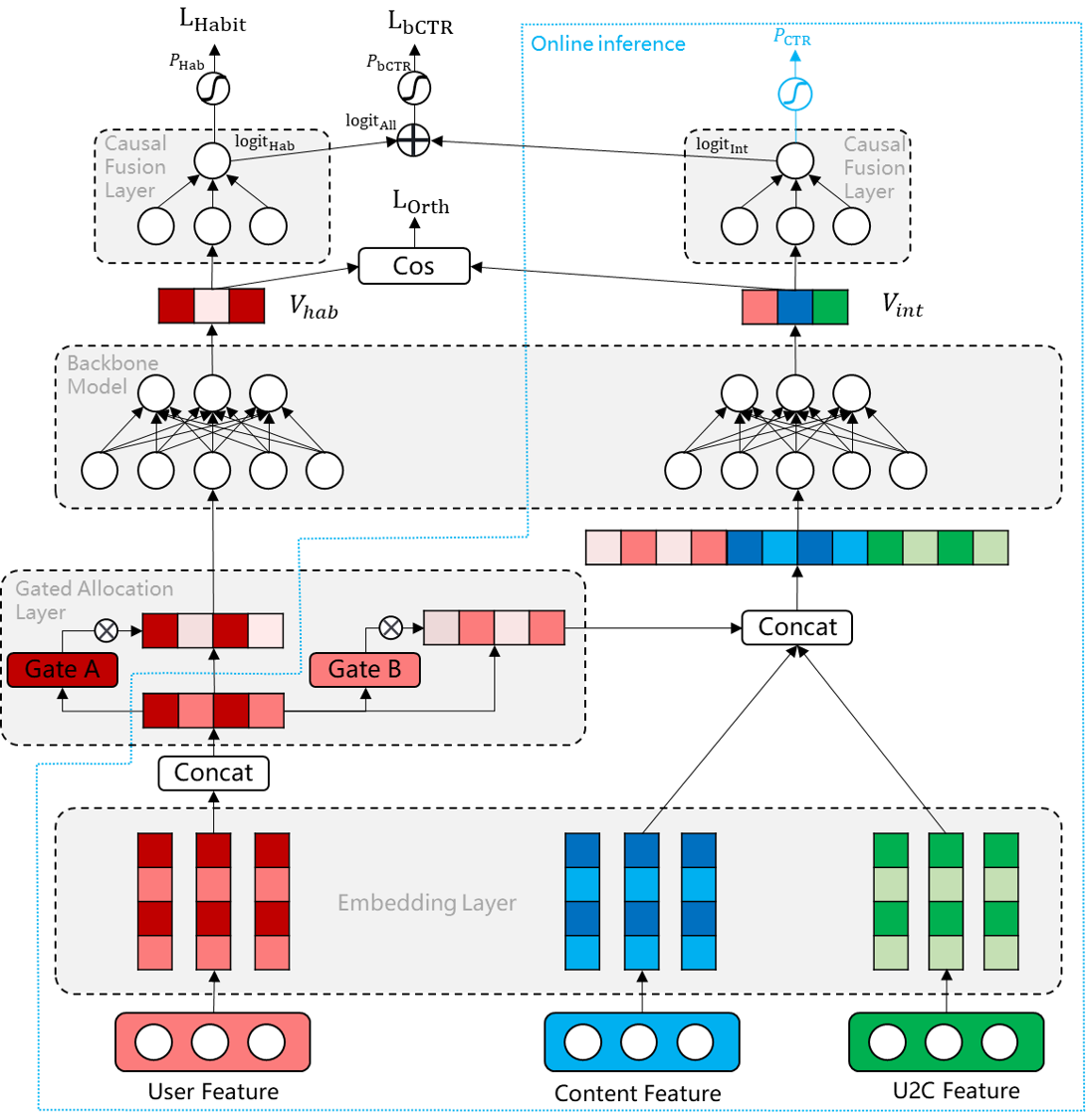}
  \caption{Model architecture of OrDA. The entire components are used for the training stage, while components within the blue box are used for the online inference stage.}
  \label{fig:OrDa}
  \Description{Model architecture of OrDA}
\end{figure}

As illustrated in Fig.~\ref{fig:OrDa}, the model architecture consists of three core components: the Gated Allocation Layer (GAL), the Backbone Model (BM), and the Causal Fusion Layer (CFL).
\subsubsection{Gated Allocation Layer (GAL)}
A potential pitfall in disentangled learning is information suppression, where the orthogonality constraint inadvertently erases generic user signals from shared embeddings. To address this, we introduce the Gated Allocation Layer (GAL) as a learnable router, avoiding all features to flow into both towers. For an input embedding $\mathbf{e}_i \in \mathbb{R}^d$, GAL computes a routing score $gate_i \in [0, 1]$ using a sigmoid-activated gating network:
\begin{equation}
gate_i = \sigma(\mathbf{w}_g^\top \mathbf{e}_i + b_g)
\end{equation}
The features are adaptively allocated into two distinct latent streams:
\begin{equation}
\begin{cases}
\mathbf{e}_{u,hab} = {gate_A \cdot \mathbf{e}_u}, \\
\mathbf{e}_{u,int} = {gate_B \cdot \mathbf{e}_u}
\end{cases}
\label{equation11}
\end{equation}
where $gate_A$ is used for the habit tower and $gate_B$ is used for the interest tower. Note that GAL is only used in user features because we assume user-channel access habits are only affected by user features (mentioned in Eq.~\ref{equation1}). This mechanism ensures that the orthogonality constraint only suppresses the redundant interference signals while allowing multi-faceted features (e.g., user demographics) to coexist in both manifolds through distinct projections. 

\subsubsection{Backbone Model (BM)}
The backbone model is an orthogonal dual-tower. The two feature streams are fed into the Habit Tower $\mathcal{F}_{hab}$ and the Interest Tower $\mathcal{F}_{int}$ respectively. Each tower is designed to project the filtered features into their respective latent manifolds:
\begin{equation}
\begin{cases}
\mathbf{v}_{hab} = \mathcal{F}_{hab}(\mathbf{e}_{u,hab}) \\
\mathbf{v}_{int} = \mathcal{F}_{int}(\text{concat}(\mathbf{e}_{u,int},\mathbf{e}_{c},\mathbf{e}_{u2c})), 
\end{cases}
\label{equation12}
\end{equation}
where $\mathbf{e}_{u,hab}$ and $\mathbf{e}_{u,int}$ is shown in Eq.~\ref{equation11}. $\mathbf{e}_{c}$ and  $\mathbf{e}_{u2c}$ are the embedding of the content features and the user2content features respectively. $\mathcal{F}_{hab}$ only processes user features. 

In order to the disentanglement theorized in our causal graphs, we enforce a orthogonal constraint (Eq.~\ref{equation7}) to ensure that $\mathbf{v}_{int} \perp \mathbf{v}_{hab}$ in the vector space, effectively "pushing" the content-driven and habit-driven signals into non-overlapping dimensions.
\subsubsection{Causal Fusion Layer (CFL)}
The final stage is the structural combination of the two towers. Following our additive formulation of causal graphs (Eq.~\ref{eq:equation4}), the logits of both towers are summed via the $\oplus$ operator:
\begin{equation}
logit_{all} = logit_{hab} \oplus  logit_{int} =  \text{MLP}_{CFL}(\mathbf{v}_{hab}) + \text{MLP}_{CFL}(\mathbf{v}_{int})
 \label{equation13}
\end{equation}
where the operator $\oplus$ denotes the joint logit fusion, representing the additive interaction between the user's intrinsic interest and their navigational habit. Our $\oplus$ operator explicitly acknowledges that the click is a superposition of two independent semantic forces. $logit_{int}$ captures the "pull" from content relevance, while $logit_{hab}$ captures the "push" from the user's ingrained access patterns on the homepage. Note that $\text{MLP}_{CFL}(\cdot)$ in Eq.~\ref{equation13} is a Multi-Layer Perceptron (MLP) as the mapping function from the vector space to the logit space.
To ensure the isolation and additivity between the two in the logit space, we use $linear$ as the activation function of the causal fusion layer.

The combined logit is then passed through a sigmoid function to produce the final click probability $P(\hat{y})$: 
\begin{equation}
P(\hat{y})=\sigma(logit_{all})
 \label{equation14}
\end{equation}
During the training stage, both $logit_{int}$ and $logit_{hab}$ are optimized simultaneously to fit the observed click labels. During the inference stage, the $\oplus$ operator allows for the seamless removal of the habit component by setting $logit_{hab} = 0$ (Eq.~\ref{equation9}), thereby yielding a purified interest score.

\section{Experiments}
\subsection{Experimental Setup}
\subsubsection{Datasets}
Production Dataset: A large-scale dataset from Zhima homepage marketing block (15-day collection), containing over 30.2 million users, 108.9 million training samples and 7.9 million evaluation samples.
\subsubsection{Baselines}
We compared with the standard model \textbf{BASE} (w/o debiasing components) and multiple SOTA debiasing methods: multi-task model \textbf{ESMM}\cite{ESMM}, propensity-based models \textbf{USD}\cite{USD} and \textbf{Multi-IPW}\cite{MultiIPW}, factorization model \textbf{PAL}\cite{PAL}.
\subsubsection{Implementation Details} We use a Multi-Layer Perceptron (MLP) as $\mathcal{F}_{hab}$ and a MaskNet\cite{MaskNet} as  $\mathcal{F}_{int}$ (Eq. ~\ref{equation12}), Adam optimizer (lr=0.0001,
batch size=2048), loss weights $\alpha$ = $\beta$ = 1 (Equation ~\ref{equation8}). Due to class imbalance in invalid exposure, we use GAUC \cite{GAUC} as primary metric: $\text{GAUC} = \frac{\sum_{u \in \mathcal{U}} w_u \times \text{AUC}_u}{\sum_{u \in \mathcal{U}} w_u}$, where $w_u$ = 1, all user group, cold-start user group (click<=1) and active user group (click>1)  correspond to
$GAUC_{all}$, $GAUC_{cold}$, $GAUC_{active}$.

\subsection{Overall Performance Comparison and Ablation Study} 
As shown in Table~\ref{tab:Performance}, \textbf{BASE} model exhibits higher $GAUC_{active}$ and lower $GAUC_{cold}$ because the model tends to "memorize" the active users' long-term, predictable access habits as strong positive signals (e.g., the users always clicking the first slot upon entering the app) but does not understand the user-content interests.
\textbf{Debiasing} models get higher $GAUC_{cold}$ performances since they tend to strip away the "habitual noise" and \textbf{OrDA} achieves the best performance both on $GAUC_{cold}$ and $GAUC_{all}$.

\begin{table}
  \caption{Performance comparison on production datasets.}
  \label{tab:Performance}
  \begin{tabular}{cccl}
    \toprule
    Model& $GAUC_{all}$& $GAUC_{cold}$ & $GAUC_{active}$\\
    \midrule
    BASE & 0.6122 & 0.6110 &  \textbf{0.6497} \\
    ESMM & 0.6299 & 0.6297 &  0.6366 \\
    USD & 0.6241 & 0.6236 & 0.6401 \\
    Multi-IPW & 0.6315 & 0.6314 & 0.6360 \\
    PAL & 0.6314 & 0.6313 & 0.6345 \\
    \midrule
    \textbf{OrDA} & \textbf{0.6412} & \textbf{0.6416} & 0.6304 \\
  \bottomrule
\end{tabular}
\end{table}

We compare OrDA with several variants as shown in Table~\ref{tab:Ablation} to investigate the impact of the core components.
\textbf{w/o GAL} (replace GAL with a standard shared-bottom) has a performance decay, which we attribute to the information evaporation effect caused by strict orthogonality, forcing all user feature embeddings to zero or random noise. Other ablated variants: \textbf{w/o HL} (remove the habit auxiliary loss function, $\alpha= 0$), \textbf{w/o OL} (remove the cosine regularization, $\beta = 0$), \textbf{w/o doC} (use the raw $logit_{int} \oplus logit_{hab}$ for ranking) face decreases of 1.06\%, 1.54\%, and 2.72\% in $GAUC_{all}$, respectively.
\begin{table}
  \caption{Ablation Study of OrDA}
  \label{tab:Ablation}
  \begin{tabular}{cccl}
    \toprule
    Model& $GAUC_{all}$& $GAUC_{cold}$ & $GAUC_{active}$\\
    \midrule
    w/o GAL & 0.6099 & 0.6108 & 0.5977 \\
    w/o HL & 0.6344 & 0.6348 & 0.6234 \\
    w/o OL & 0.6313 & 0.6311 & \textbf{0.6393} \\
    w/o doC & 0.6237 & 0.6235 & 0.6331 \\
    \midrule
    \textbf{OrDA} & \textbf{0.6412} & \textbf{0.6416} & 0.6304 \\
  \bottomrule
\end{tabular}
\end{table}

\subsection{Latent Space Visualization}
\begin{figure}[h]
    \centering
    \includegraphics[width=0.8\linewidth]{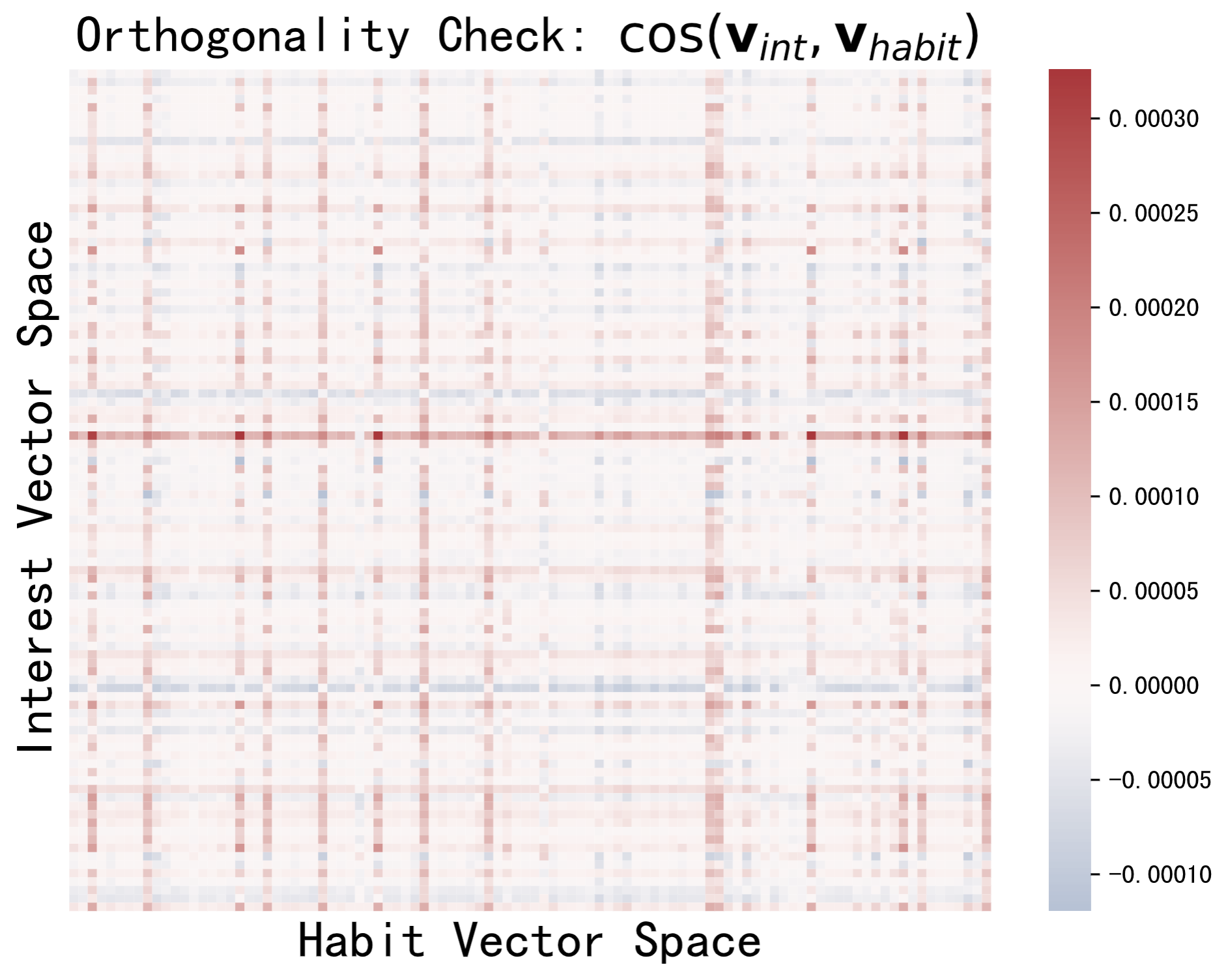}
    \caption{Visualizing the orthogonality between interest and habit vectors. 
    The color of the heatmap is predominantly white/light, indicating that the cosine similarity values are concentrated around 0.
    }
    \label{fig:Orthogonality_Check_OrDA}
\end{figure}

To quantitatively and visually assess the degree of disentanglement, we compute the cosine similarity matrix between the interest vectors $\mathbf{v}_{int}$ and the habit vectors $\mathbf{v}_{habit}$ across a random subset of the evaluation set. As illustrated in Fig.~\ref{fig:Orthogonality_Check_OrDA}, the similarity heatmap exhibits a near-zero distribution across both the diagonal (intra-sample) and off-diagonal (inter-sample) elements, demonstrating the two vector spaces are constrained to be mutually orthogonal, thereby confirming that the disentanglement is both sample-wise consistent and globally robust.

\subsection{Online A/B Test}
We run a week-long online A/B test on Zhima homepage marketing blocks: Zhima’s rent-floor, where \textbf{OrDA} outperforms the baseline model (\textbf{BASE}, see Sec.4.1.2) with UCTR increased 5.64\%. This substantial gain demonstrates the industrial validity of OrDA.

\section{Conclusion}
We propose the OrDA framework designed to disentangle user-content intrinsic interest from user-channel access habit on homepage recommendations. Extensive experiments on large-scale industrial datasets and the online A/B test demonstrate its effectiveness, effectively recovering purified interest scores. Notably, OrDA has been fully deployed in Zhima homepage marketing blocks: Zhima’s rent-floor recommendation.


\bibliographystyle{ACM-Reference-Format}
\bibliography{main}

\appendix

\end{document}